\documentclass[10pt,twocolumn,letterpaper]{article}

\usepackage{cvpr}              % To produce the CAMERA-READY version
\usepackage{multirow}
\usepackage{makecell}
\usepackage{bm}  

\definecolor{cvprblue}{rgb}{0.21,0.49,0.74}
\usepackage[pagebackref,breaklinks,colorlinks,allcolors=cvprblue]{hyperref}

\newcommand{\myPara}[1]{\vspace{6pt}\noindent\textbf{#1}}

\def\paperID{5133} % *** Enter the Paper ID here
\def\confName{CVPR}
\def\confYear{2025}

\title{KAC: Kolmogorov-Arnold Classifier for Continual Learning}

\author{Yusong Hu$^1$, Zichen Liang$^1$, Fei Yang$^{1,2}$, Qibin Hou$^{1,2}$, Xialei Liu$^{1,2}$\thanks{Corresponding author.}, Ming-Ming Cheng$^{1,2}$\\ 
$^1$VCIP, CS, Nankai University \qquad $^2$NKIARI, Shenzhen Futian\\
{\tt\small \{ethanhu, liangzc\}@mail.nankai.edu.cn} \\
{\tt\small\{feiyang, houqb, xialei, cmm\}@nankai.edu.cn}
}

\begin{document}
\maketitle
\begin{abstract}
Continual learning requires models to train continuously across consecutive tasks without forgetting. Most existing methods utilize linear classifiers, which struggle to maintain a stable classification space while learning new tasks. Inspired by the success of Kolmogorov-Arnold Networks (KAN) in preserving learning stability during simple continual regression tasks, we set out to explore their potential in more complex continual learning scenarios. In this paper, we introduce the Kolmogorov-Arnold Classifier (KAC), a novel classifier developed for continual learning based on the KAN structure. We delve into the impact of KAN's spline functions and introduce Radial Basis Functions (RBF) for improved compatibility with continual learning. We replace linear classifiers with KAC in several recent approaches and conduct experiments across various continual learning benchmarks, all of which demonstrate performance improvements, highlighting the effectiveness and robustness of KAC in continual learning. The code is available at \url{https://github.com/Ethanhuhuhu/KAC}.
\end{abstract}    
\section{Introduction}

Deep learning models are typically trained on a fixed dataset in a single session, achieving impressive performance on various static tasks. In contrast, real-world scenarios continuously evolve, necessitating models that can learn incrementally from a data stream. 
However, in such scenarios, these models often encounter a significant challenge, known as catastrophic forgetting~\cite{FRENCH1999128}. 
Continual learning~\cite{9349197,BELOUADAH202138,PARISI201954,10.1145/776985.776986} investigates how to effectively train models in such dynamic environments with sequential data exposure, aiming to adapt and avoid forgetting over time.

\begin{figure}[t]
    \centering
    \begin{subfigure}[b]{0.98\linewidth}
        \centering
        \includegraphics[width=\textwidth]{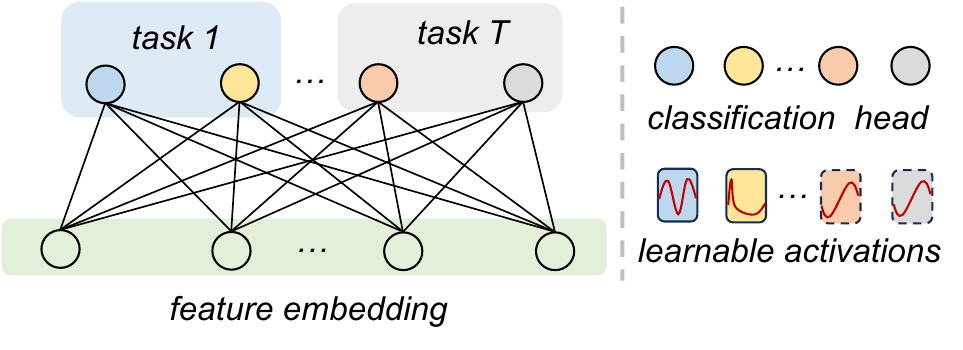}
        \caption{Linear Classifier}
        \hspace{10pt}
        \label{subfig:a}
    \end{subfigure}
        \begin{subfigure}[b]{0.98\linewidth}
        \centering
        \includegraphics[width=\textwidth]{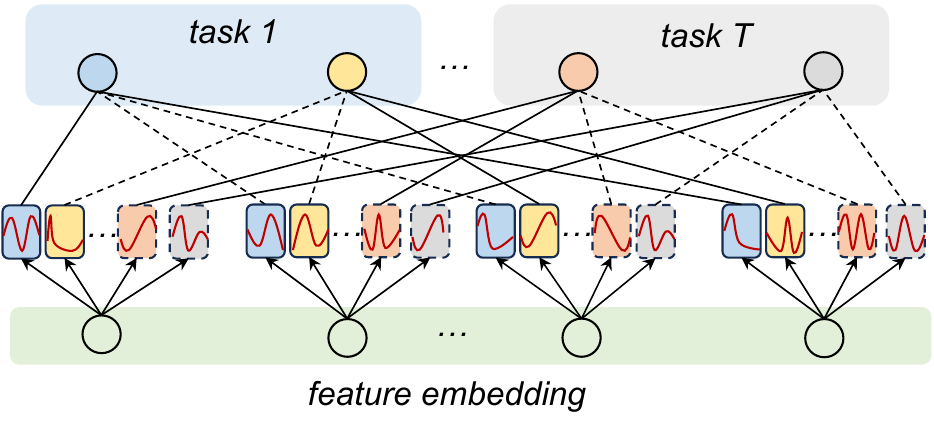}
        \caption{Kolmogorov-Arnold Classifier}
        \label{subfig:b}
    \end{subfigure}
    %传统的分类器在所有任务当中都均等的激活每一个权重，而我们的Kolmogorov-Arnold Classifier为每一个通道在每一个类别上学习一个class-specific的learnable activation，最大程度减小了无关权重变化导致的forgetting。
    \caption{Brief comparison between conventional linear classifier and our Kolmogorov-Arnold classifier. The solid lines represent activated weights, while the dashed ones represent suppressed weights. (a) Conventional linear classifiers activate each weight equally across all tasks, resulting in irrelevant weights being equally updated in the new task.. (b) our Kolmogorov-Arnold Classifier learns class-specific learnable activations for each channel across all categories, minimizing forgetting caused by irrelevant weight changes.}
    \label{fig: comparison}
\end{figure}

Class incremental learning (CIL)~\cite{rebuffi2017icarl}, as a key challenge in continual learning, has garnered extensive research interest. It involves the continuous introduction of new classes with ongoing tasks, requiring the model to conduct classification on all encountered classes after training on new tasks. Most CIL methods retain exemplars and employ techniques, such as knowledge distillation~\cite{rebuffi2017icarl,douillard2020podnet,wen2024class} or dynamic architectures~\cite{chen2023dynamic,douillard2022dytox,yan2021dynamically,kim2024cross}, to mitigate forgetting. 
With the development of pre-trained models, numerous studies~\cite{mcdonnell2024ranpac,zhang2023slca} have attempted to explore the applications of pre-trained models in CIL, achieving impressive results.
Among these, prompt-based approaches~\cite{wang2022learning,wang2022dualprompt,smith2023coda,gao2024consistent} have attracted considerable attention.

For existing methods, some \cite{mcdonnell2024ranpac,goswami2024fecam,Yu_2020_CVPR} focus on feature space design through carefully crafted classifiers and training or inference strategies, achieving excellent performance. 
These studies demonstrate that a well-structured feature space can effectively mitigate the forgetting issue in that a stable distribution is crucial for continual classification tasks while the design of classifiers is essential for constructing the feature space and reducing forgetting in continuous tasks. 
However, most existing approaches~\cite{gao2024consistent,zhou2024expandable,smith2023coda} use linear classifiers or nearest class mean classifiers (NCM)~\cite{rebuffi2017icarl}, with limited research focused on developing a specific classifier for CIL to effectively mitigate catastrophic forgetting, which warrants further study.
%
% Therefore, designing an efficient classifier that replaces the existing simple classifiers and enhances current approaches will significantly advance the development of CIL.

% \begin{figure*}[t]
%     \centering
%     \begin{subfigure}[b]{0.38\textwidth}
%         \centering
%         \includegraphics[width=\textwidth]{images/Fig.1.a.pdf}
%         \caption{Linear Classifier}
%         \label{subfig:a}
%     \end{subfigure}
%     \begin{subfigure}[b]{0.50 \textwidth}
%         \centering
%         \includegraphics[width=\textwidth]{images/Fig.1.b.pdf}
%         \caption{Kolmogorov-Arnold Classifier}
%         \label{subfig:b}
%     \end{subfigure}
%     %传统的分类器在所有任务当中都均等的激活每一个权重，而我们的Kolmogorov-Arnold Classifier为每一个通道在每一个类别上学习一个class-specific的learnable activation，最大程度减小了无关权重变化导致的forgetting。
%     \caption{Brief comparison between conventional linear classifier and our Kolmogorov-Arnold classifier. The solid lines represent activated weights, while the dashed ones represent suppressed weights. Conventional linear classifiers activate each weight equally across all tasks, whereas our Kolmogorov-Arnold Classifier learns class-specific learnable activations for each channel across all categories, minimizing forgetting caused by irrelevant weight changes.}
%     \label{fig: comparison}
% \end{figure*}

Recently, a novel architecture, Kolmogorov–Arnold Networks (KAN)~\cite{liu2024kan} , has been proposed, demonstrating natural effectiveness in continual learning. The authors compare KAN with Multi-Layer Perceptrons (MLP)~\cite{hornik1989multilayer} on a toy continual 1D regression problem, which requires the model to fit 5 Gaussian peaks sequentially. KAN exhibits superior performance, effectively mitigating catastrophic forgetting, attributed to the locality of splines and inherent local plasticity. This locality allows KAN to identify relevant regions for re-organization while maintaining stability in other areas during sequential tasks~\cite{liu2024kan}. These findings motivate us to explore the applications of KAN in more challenging CIL tasks.

In this paper, we introduce the Kolmogorov-Arnold Classifier (KAC), a plug-and-play classifier for Continual Learning built upon the KAN architecture. Leveraging the Kolmogorov-Arnold representation theorem~\cite{kolmogorov1961representation}, we integrate learnable activation functions on the edges of the classifier. 
% %
% We find that the conventional KAN with B-spline functions struggles with high-dimensional data, leading to inadequate model plasticity, which may weaken the models' plasticity when directly introduced as a classifier.
%
We find that the conventional KAN with B-spline functions struggles with high-dimensional data, resulting in insufficient model plasticity, which may reduce the model’s adaptability when directly employed as a classifier. This limitation forces models to undergo excessive updates when learning new tasks, resulting in significant forgetting. 

To address this, we explore spline functions and identify Radial Basis Functions (RBF) as an effective alternative for continual learning. By utilizing RBF in our KAC, we enhance the model's ability to adapt CIL while minimizing forgetting.
Thanks to these learnable spline activations, the KAC allows the model to select specific activation ranges of interest for each channel while preserving the distribution of other parts, and RBF makes it more compatible with CIL. 
As shown in Fig.~\ref{subfig:b}, these learnable activations help the model select interesting parts of each channel and activate them for determination rather than activating all edges like a simple linear classifier in Fig.~\ref{subfig:a}.
This brings notable benefits to class incremental learning.
When new tasks arrive, the learnable activation functions assist the model in selecting relevant parts of each channel for updating. This prevents the drift of irrelevant features during the training process for the new tasks. Meanwhile, the deactivated portions of the old tasks remain unaffected by these updates, reducing the forgetting of old tasks.

To demonstrate the superiority of KAC, we conduct experiments on several prompt-based continual learning approaches, which are built upon a pre-trained backbone where the classifiers play a key role in these approaches. The models employing our method achieve significant improvement across various CIL scenarios on multiple datasets by simply replacing the linear classifier with our KAC without making any other modifications or hyperparameter adjustments. Additionally, experiments conducted in the Domain Incremental Learning (DIL)~\cite{wang2022s} setting reveal that our method can also improve performance, demonstrating its effectiveness and robustness. 

Our main contributions can be summarised as follows:
\begin{itemize}

    \item We explore the application of Kolmogorov-Arnold Networks (KAN) in continual learning and analyze its weaknesses when employed in continual learning and how to enhance its compatibility with such tasks.

    \item We introduce the Kolmogorov-Arnold Classifier (KAC), a novel continual classifier based on the KAN structure with Radial Basis Functions (RBF) as its basis functions. KAC enhances the stability and plasticity of CIL approaches.

    \item We integrate our KAC into various approaches and validate their performance across multiple continual learning benchmarks. The results demonstrate that KAC can effectively reduce forgetting in these methods.
\end{itemize}

\section{Related Work}

\myPara{Class Incremental Learning} aims to learn a sequence of classification tasks sequentially, where the number of classes increases with each task. The primary challenge in it is catastrophic forgetting\citep{mccloskey1989catastrophic}. Several studies work on it and they can be broadly categorized into three main strategies: regularization-based, structure-based, and replay-based methods. Regularization-based methods reduce forgetting by employing knowledge distillation techniques\citep{wen2024class, yang2022uncertainty, douillard2020podnet} or imposing constraints on key model parameters\citep{kang2022class, kirkpatrick2017overcoming,liu2022long}. Structure-based methods\citep{chen2023dynamic, wang2022beef, douillard2022dytox,10681473} mitigate forgetting through dynamic network architectures. 
Replay-based methods retain a small portion of old data\citep{pmlr-v202-jeeveswaran23a, rebuffi2017icarl} or use auxiliary models\citep{kim2024sddgr, gao2023ddgr, shin2017continual} to generate synthetic data, which are combined with new-class data to update the model.

\myPara{CIL with Pre-trained Models} have demonstrated their competitive performance in Class Incremental Learning due to their strong transferability. Techniques such as LAE \citep{gao2023unified} and SLCA \citep{zhang2023slca} enhance model adaptation through EMA-based updates and dynamic classifier adjustments. RanPAC \citep{mcdonnell2024ranpac} employs random projection to improve continual learning, while EASE \citep{zhou2024expandable} focuses on optimizing task-specific, expandable adapters to enhance knowledge retention. Benefiting from parameter-efficient tuning in NLP, prompt-based methods have achieved promising results in Class Incremental Learning. These approaches utilize adaptive prompts to guide frozen transformer models, facilitating efficient task-specific learning without modifying encoder parameters. Techniques like L2P \citep{wang2022learning}, DualPrompt \citep{wang2022dualprompt}, S-Prompts \citep{wang2022s}, CODA-Prompt \citep{smith2023coda}, HiDe-Prompt \citep{wang2024hierarchical}, and CPrompt \citep{Gao_2024_CVPR} introduce diverse prompt selection strategies to improve task learning, knowledge retention, and model robustness.

\myPara{Kolmogorov-Arnold Networks} (KAN)~\citep{li2024kolmogorov} is a novel network architecture based on the Kolmogorov-Arnold representation theorem~\citep{kolmogorov1961representation}. It represents multivariate functions as combinations of multiple univariate functions and uses nonlinear spline functions for approximation. Some explorations focus on how to apply KAN to solve scientific problems~\citep{koenig2024kan,bozorgasl2024wav,howard2024finite}, while others seek various basis functions to enhance performance~\citep{aghaei2024fkan,bozorgasl2024wav,li2024kolmogorov}. Many works~\citep{bresson2024kagnns,de2024kolmogorov,genet2024tkan,10564665} apply KAN across various fields and investigate how to effectively leverage its advantages in these domains. These studies encourage us to explore the application of KAN in continual learning.

\section{Method}

\subsection{Preliminaries}
\label{sec:preliminaries}

\myPara{Class Incremental Learning.}
In Class Incremental Learning (CIL), a model needs to learn classes step by step.
At each step \( t \), the model needs to learn the classes specific to that step, denoted as \(\mathcal{Y}_t\), with only access to the current dataset \( D_t = \{(\bm{x}_t^{i}, y_t^{i})\}_{i=1}^{n_t} \), where \( \bm{x}_t^{i} \) represents an input image and \( y_t^{i} \) is its corresponding label.
% A pivotal challenge lies in the model's ability to assimilate new categories without compromising its retention of previously learned ones.
A key challenge in CIL is how to maintain the stability of the model to avoid catastrophic forgetting~\citep{FRENCH1999128} while learning new tasks.
% enhance the model's plasticity for learning new classes while maintaining stability to prevent forgetting old ones.
% Except for the widely studied Class-Incremental Learning (CIL) scenario, there are also Task-Incremental Learning (TIL)~\citep{oren2021defense,feng2019challenges}, where task labels are available during testing, and the more challenging Domain-Incremental Learning (DIL)~\citep{wang2022s,li2024coleclip}, where the model lacks task labels and faces varying input distributions with consistent classes.
With a model consisting of a backbone $F$, and a classifier $h \in \mathbb{R}^{n \times C}$, where $n$ denotes the embedding dimension and $C$ represents the total number of learned classes, the model is tasked with predicting the class label $y = h(F(\bm{x})) \in \mathcal{Y}$ for test samples from new classes as well as samples from previously encountered tasks.
% For prompt-based incremental learning, the backbone function $f$ benefits from pre-training on a large corpus of external data, which is separate from each dataset $D_t$.
% In this pre-training scenario, the model operates without rehearsal during the continual learning process, meaning that each element of $D_t$ is only accessible while executing task $t$.

\myPara{Kolmogorov–Arnold Networks.}
Kolmogorov–Arnold Network (KAN)~\citep{liu2024kan} is a novel model architecture that serves as a promising alternative to multi-layer perceptrons (MLPs)~\citep{haykin1998neural,hornik1989multilayer}. While MLPs rely on the Universal Approximation Theorem (UAT)~\citep{hornik1989multilayer}, KANs are inspired by the Kolmogorov-Arnold representation Theorem (KAT)~\citep{kolmogorov1961representation}. KAT posits that any multivariate continuous function $f(\bm{x})$ defined on a bounded domain can be expressed as a finite composition of univariate continuous functions through addition. The Kolmogorov-Arnold representation theorem can be written as:

\begin{equation}
    f(\bm x) = f({x}_1, {x}_2, ..., {x}_n) = \sum\limits_{q=1}^{2n+1}\Phi_q \Big(\sum\limits^{n}_{p=1}\phi_{q,p}( x_p)\Big),
\end{equation}

\noindent in which $\Phi_q$ and $\phi_{q,p}$ are univariate functions for each variable. KAN parametrizes the $\phi_{q,p}$ and $\Phi_q$ as B-spline curves, with learnable coefficients of local B-spline basis functions $B(\bm x)$~\citep{qin1998general}. In practice, a residual connection, consisting of a linear function with activation $b(\bm x)=silu(\bm x)=\bm x/(1+e^{-\bm x})$, is linearly combined with the B-spline curve $spline(\bm x)=\sum_i\omega_iB_i(\bm x)$ to form the final $\phi$:

\begin{equation}
    \phi(x)=\omega_bb(\bm x) + \omega_sspline(\bm x),
\end{equation}

\noindent where the $\omega_b$ and $\omega_s$ represent the linear functions that control the overall magnitude of the activation function. Consequently, a KAN layer can be expressed as:

\begin{equation}
    \begin{aligned}
        % x_{l+1} = \Phi_l(x_l),
        \bm{x}_{l+1} &= \underbrace{\begin{pmatrix} \phi_{l,1,1}(.) & \phi_{l,1,2}(.) & \cdots & \phi_{l,1,n_l}(.) 
        \\\phi_{l,2,1}(.) & \phi_{l,2,2}(.) & \cdots & \phi_{l,2,n_l}(.) 
        \\ \vdots & \vdots & \ddots & \vdots 
        \\\phi_{l,n_{l+1},1}(.) & \phi_{l,n_{l+1},2}(.) & \cdots & \phi_{l,n_{l+1},n_l}(.) 
        \end{pmatrix}}_{\Phi_l}
        \bm{x}_l.
    \end{aligned}
\end{equation}

% in which the $\Phi_l$ is defined like that:
% \begin{equation}
%     % \begin{aligned}
%         % x_{l+1} &= \Phi_l(x_l)\\
%         \begin{pmatrix} \phi_{l,1,1}(.) & \phi_{l,1,2}(.) & \cdots & \phi_{l,1,n_l}(.) 
%         \\\phi_{l,2,1}(.) & \phi_{l,2,2}(.) & \cdots & \phi_{l,2,n_l}(.) 
%         \\ \vdots & \vdots & \ddots & \vdots 
%         \\\phi_{l,n_{l+1},1}(.) & \phi_{l,n_{l+1},2}(.) & \cdots & \phi_{l,n_{l+1},n_l}(.) 
%         \end{pmatrix}.
%     % \end{aligned}
% \end{equation}

The $\bm{x}_l$ and $\bm{x}_{l+1}$ represent the input and output of a KAN layer, while $\phi_{l}$ is the 1D univariate function matrix for each layer. The KAN networks are constructed by stacking multiple KAN layers.

\subsection{Conventional KAN layer is not a good continual classifier}
\label{sec:ori_kan}
\begin{figure*}[tbp]
    \centering
    \setlength{\abovecaptionskip}{2pt}
    
    \begin{subfigure}[b]{0.32\linewidth}
        \centering
        \includegraphics[width=\linewidth]{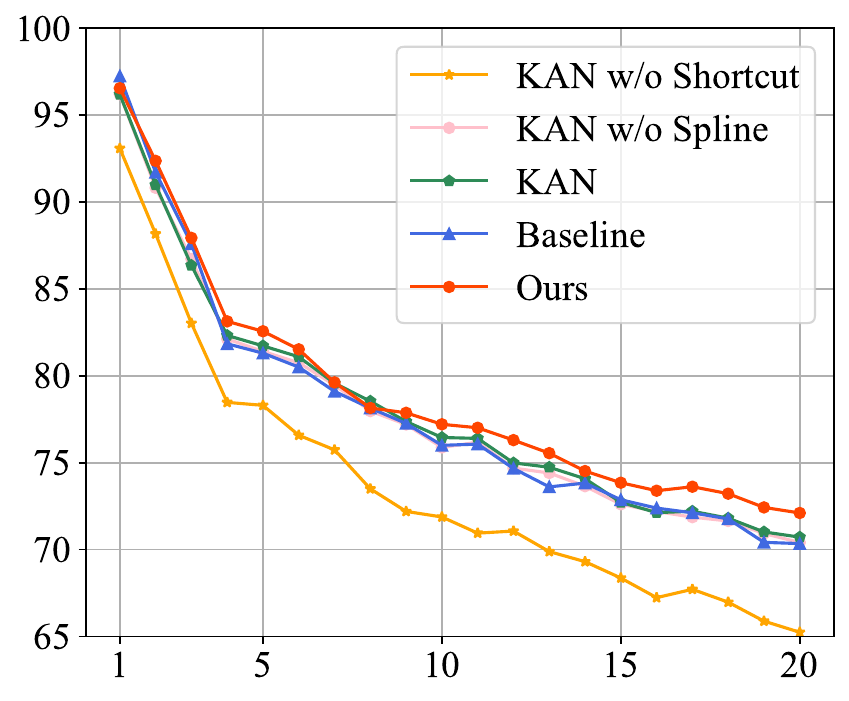}
        \caption{L2P}
    \end{subfigure}
    \hfill
    \begin{subfigure}[b]{0.32\linewidth}
        \centering
        \includegraphics[width=\linewidth]{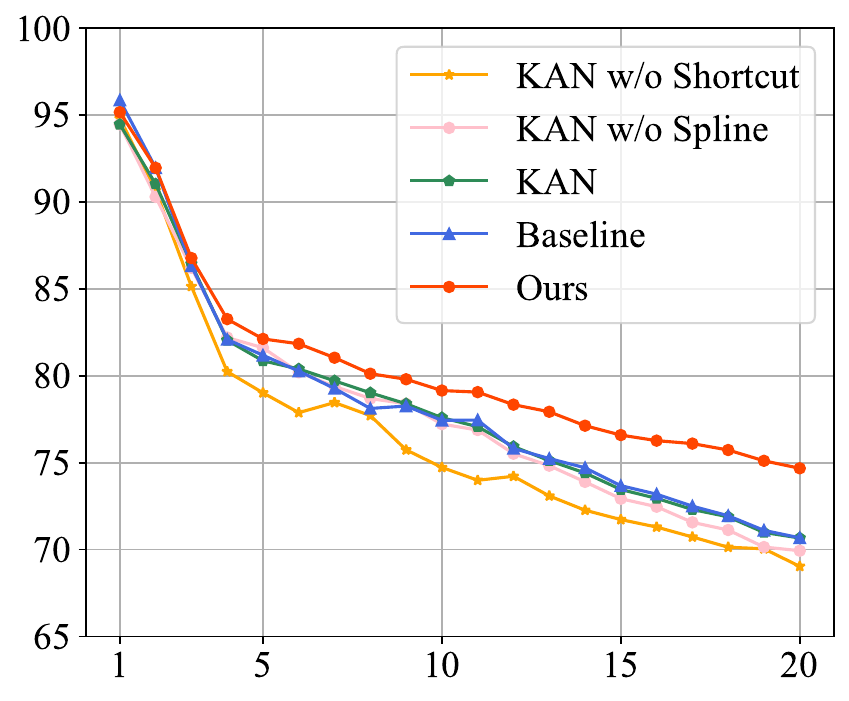}
        \caption{DualPrompt}
    \end{subfigure}
    \hfill
    \begin{subfigure}[b]{0.32\linewidth}
        \centering
        \includegraphics[width=\linewidth]{images/20-task-dual-prompt.pdf}
        \caption{CODAPrompt}
    \end{subfigure}
    %Comparison of 几种方法with不同分类器在ImageNet-R 20steps场景下的精度曲线。The x-axis represents the increasing tasks, while the y-axis shows the corresponding test accuracy at each step. While Baseline represents their performance with a conventional linear classifier, other curves are with ablated KAN classifiers and our Kolmogorov-Arnold Classifier.
    \caption{Comparison of the accuracy curves of three recent approaches with different classifiers in the ImageNet-R 20-step scenario. The x-axis represents the increasing number of tasks, while the y-axis shows the corresponding test accuracy at each step. The Baseline indicates performance with a conventional linear classifier, while the other curves represent results with ablated KAN classifiers and our Kolmogorov-Arnold Classifier.}
    \label{fig: ImageNetr}
\end{figure*}

In \cite{liu2024kan}, the authors present experimental results from a toy 1D regression task, demonstrating that the locality of splines can inherently avoid catastrophic forgetting. This insight inspires us to introduce KAN to CIL.
A straightforward way to leverage the locality of KAN is directly utilizing a KAN layer to develop a continual classifier, replacing the linear classifier in CIL methods. To achieve this, we simply replace the linear classifier $h(\bm x)$ with a KAN layer that has an input dimension of $d$ and an output dimension of $C$. We compared their performances across several baseline methods. The experimental results are shown in Fig.~\ref{fig: ImageNetr}, demonstrating that the simple substitution of replacing the linear classifier with a KAN layer does not lead to any improvement, even achieving worse performance.

We decompose the KAN layer into two parts: the residual connection $b(\bm x)$ and the B-spline curve $spline(\bm x)$ and individually replace the linear classifier with these two components to investigate why directly introducing the KAN layer increases forgetting. A surprising finding is that the B-spline functions lead to a severe performance drop across all baselines.

To understand why the B-spline curve replacing the conventional linear classifier leads to severe forgetting, we need to delve deeper into the differences between linear layers and splines. In high-dimensional complex data, spline functions encounter the curse of dimensionality (COD)~\citep{hammer1962adaptive}; as the data dimensionality increases, the model struggles with data approximation~\citep{koppen2002training,montanelli2020error,he2023optimal}. This is because splines cannot effectively model the compositional structure present in the data, while linear classifiers benefit from their fully connected structure, allowing them to learn this structure effectively~\citep{he2023deep}. Although KAN networks mitigate COD through approximation theory~\citep{liu2024kan} by stacking KAN layers, approximating high-dimensional function remains a challenging problem for a single spline layer, whereas it is relatively straightforward for conventional linear classifiers.

It is precisely the weak fitting ability of B-spline functions on high-dimensional data that leads to severe forgetting when it is introduced into CIL. In CIL, a network typically consists of a backbone $F$ that encodes images to feature embeddings and a classification head $h$, which serves as a high-dimensional projection mapping the embeddings to class probabilities. Most methods accommodate new classes by adding classifiers while sharing the backbone across all tasks. The final logits $l$ for classification are always calculated as:
\begin{equation}
    l = h(F(\bm x)), h = [h_1, h_2, \cdots, h_t].
\end{equation}

To prevent significant forgetting caused by changes in the backbone that affect the feature space, the model must maintain stable backbone parameters during training on new tasks. Consequently, many methods use regularization techniques to restrict changes in feature embeddings~\citep{li2017learning,kim2024cross,wen2024class, yang2022uncertainty}. However, due to the limited approximation capability of a single B-spline layer, the model requires more extensive updates to the backbone parameters compared to conventional linear classifiers to achieve good performance on new tasks. This extensive updating can severely disrupt the feature space, leading to pronounced forgetting.

Based on the above analysis, we believe that the weak fitting ability of a single B-spline function prevents the model from leveraging the locality of the KAN layer. Therefore, we need to enhance the spline function's fitting ability to adapt the KAN structure to CIL tasks. 
% a表明，在特定的空间当中，浅层的基于KAT的network layer能够通过设计过的basis function来break the COD when approximating high dimensional functions.这启发我们思考什么样的basis function适合CIL任务
\cite{lin2017does,lai2021kolmogorov} indicates that, in specific senses, a shallow KAT-based layer can break the COD problem when approximating high-dimensional functions through designed basis functions with particular compositional structures, motivates us to explore the types of basis functions that are compatible with CIL.

\subsection{Radial basis function is great for class incremental learning}
\label{sec:rbf}

Several studies~\citep{mcdonnell2024ranpac,Zhuang_2023_CVPR,Yu_2020_CVPR} assume that the classification space follows a Gaussian space and develop approaches based on this premise, achieving excellent performance. It suggests that building a Gaussian classification space can help models effectively learn new tasks while combating catastrophic forgetting. Can we find a kind of basis function in this sense that allows a KAT-based layer function as a continual classifier, addressing the COD problem and benefiting CIL? The answer is yes! 

\begin{figure*}[t]
    \centering
    \setlength{\abovecaptionskip}{2pt}
    \includegraphics[width=1\textwidth]{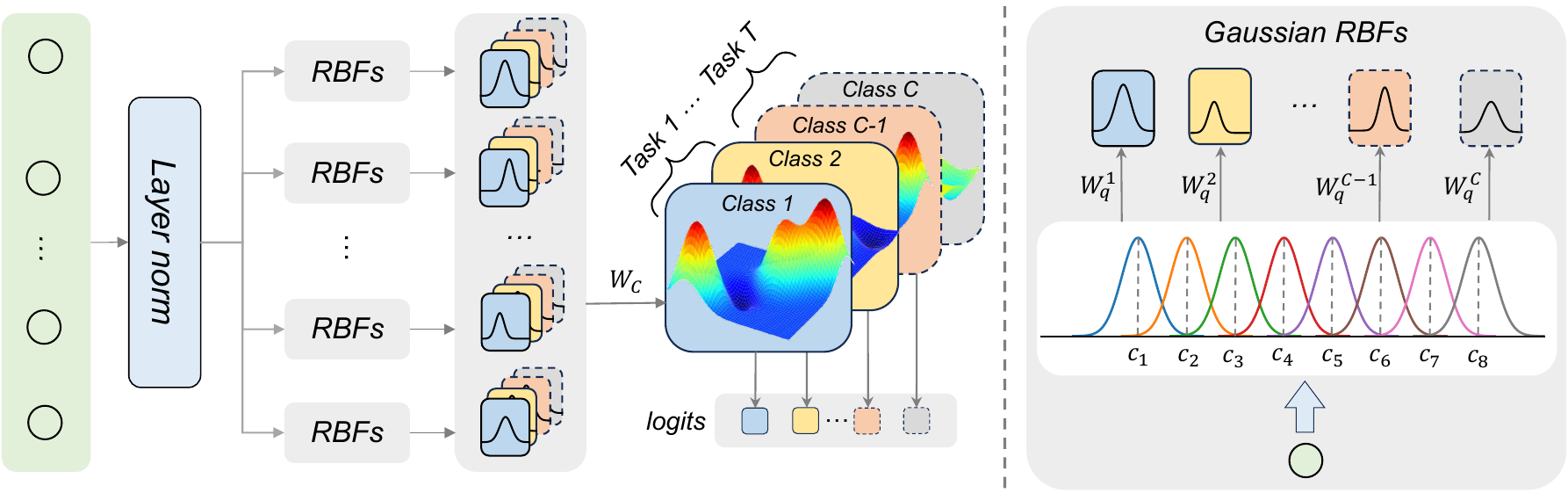}
    \vspace{0.5pt}
    \caption{An overview of the pipeline of the proposed Kolmogorov-Arnold Classifier. For the input feature embeddings, we first normalize them using a layer normalization, then pass them through a set of RBFs that activate them to learnable Gaussian distributions. Finally, we weight all channels with $W_C$ to obtain the decision space for each class. The right side shows the process of Gaussian RBFs, which map univariate variables to different Gaussian distributions centered at various points and weight these distributions with $W_q^c$ to derive the final activation distribution for each channel across all classes. The output logits are sampled based on the channel values within the distribution of each class.
    As tasks increase, new classes can be accommodated by simply expanding $W_C$.}
    \label{fig:pipeline}
\end{figure*}

FastKAN~\citep{li2024kolmogorov} proves that the B-splines basis function in KAN~\citep{liu2024kan} can be well replaced by Radial Basis Functions (RBF)~\citep{buhmann2000radial,orr1996introduction}. We find this substitution brings more benefits to CIL when KAN is introduced as a continual classifier as shown later. A KAN layer with RBF is represented as:
\begin{equation}
    f(\bm x) = \sum\limits_{p=1}^{n}\Phi_p
    % \sum\limits^{n}_{p=1}
    \sum\limits^{N}_{i=1}   \omega_{p,i}\phi(|| {x}_p - c_i ||),
\end{equation}
where $c_i$ represents a series of center points evenly distributed within a specific range, with $N$ denoting the total number of $c_i$. And $\phi(.)$ is an RBF served as the basis functions whose value solely depends on the distance between input $x_p$ and center point $c_i$. The term $\omega_{p,i}$ denotes the weight for each $\phi$. A Gaussian function with covariance $\sigma_i$ can be chosen as $\phi$ while it's defined as:
\begin{equation}
\label{eq:theta}
    \phi(||{x}_p - c_i||) = \mathrm{exp} \Big(-\frac{{({x}_p - c_i)}^2}{2\sigma_i^2}\Big).
\end{equation}

While introducing the Gaussian RBF function as the basis function of KAN demonstrates faster evaluation speeds and enhanced performance, as shown in \cite{li2024kolmogorov}, an inherent Gaussian structure is also established with it, which can serve as an effective compositional structure for CIL scenarios. 
%%%%%%%%%%%%%%%%%%%%%%%%%%%%%%%%%%%%%%%%%%%%%%%%%%%%%%%%%%%%%%%
With a series of Gaussian distributions $\mathcal{N}$ centered at $c = [c_1, c_2, \cdots, c_N]$, the activation function for each dimension is formed by combining $N$ Gaussian distributions, and the distribution of each dimension can be represented as a Gaussian mixture model:

\begin{equation}
    \begin{aligned}
        \sum\limits^{N}_{i=1} \omega_{p,i}\phi(\| {x}_p - c_i \|)
        &\sim \sum\limits^{N}_{i=1} \omega_{p,i}\mathcal{N}(c_i, \sigma_i^2) \\
        &= \omega_{p,1}\mathcal{N}(c_1, \sigma_1^2) + \omega_{p,2}\mathcal{N}(c_2, \sigma_2^2) \\ 
        &\quad + \cdots + \omega_{p,N}\mathcal{N}(c_N, \sigma_N^2)
    \end{aligned}
\end{equation}

This mixture formulation preserves the multi-modal characteristics of the original Gaussian components. The final prediction for each class is then expressed as a dimension-wise weighted combination of these Gaussian mixtures:

\begin{equation}
    f(\bm{x}) = \sum\limits^{n}_{p=1} \Phi_p \left[ \sum\limits^{N}_{i=1} \omega_{p,i} \exp\left(-\frac{(x_p - c_i)^2}{2\sigma_i^2}\right) \right]
\end{equation}

We can easily derive that, thanks to the introduction of Gaussian RBF functions, the features of $p$th dimension in the KAN layer, after the activation function, follow a Gaussian mixture distribution.
% with mean $\mu_p=\sum^{N}_{i=1}\omega_{p,i}c_i$ and variance $\sigma_p=\sum^N_{i=1}\omega_{p,i}^2\sigma_i^2$. This results in the final prediction for each class being represented as the sum of a set of Gaussian distributions, represented as:
% \begin{equation}
%     f(\bm x) = \sum\limits^{n}_{p=1}\Phi_p\cdot \mathrm{exp}\Big(-\frac{( {x}_p - \mu_p)^2}{2\sigma_p^2}\Big).
%     % , \mu_p=\sum^{N}_{i=1}\omega_{p,i}c_i, \sigma_p=\sum^N_{i=1}\omega_{p,i}^2\sigma_i^2.
% \end{equation}
When we simply define $\Phi_p$ as a learnable weight for each dimension, it is evident that the resulting function form conforms to the Gaussian Process (GP) with first-order additive kernels defined in \cite{duvenaud2011additive}. This structure is consistently easy to fit for classification tasks and possesses a strong long-range structure to effectively address the COD problem when approximating high-dimensional functions~\citep{duvenaud2011additive}. With functions like this serving as the basis functions for continual classifiers, it not only projects each channel of the feature into a Gaussian space but also allows the model to select an interested range for each channel tailored to different classes.

%%%%%%%%%%%%%%%%%%%%%%%%%%%%%%%%%%%%%%%%%%%%%%%%%%%%%%%%%%%%%%%

% \textcolor{red}{revision}
%%%%%%%%%%%%%%%%%%%%%%%%%%%%%%%%%%%%%%%%%%%%%%%%%%%%%%%%%%%%%%%
\newcommand{\changeval}[1]{\textcolor{red}{\ \tiny( #1\ )}}
\newcommand{\lessval}[1]{\textcolor{blue}{\ \tiny( #1\ )}}

\begin{table*}[t]
    \centering
    \setlength{\tabcolsep}{7pt}
    \small
    %我们在imagenrtr的5,10,20,40 steps的CIL场景中比较了多种方法在linear classifier （baseline）和加入我们的KAC的结果，结果表明我们的方法均为他们带来了提升，尤其是在长序列任务当中。
    \caption{Results on ImageNet-R dataset. We report the average incremental accuracy and the last accuracy on CIL scenarios of 5, 10, 20, and 40 steps and make comparisons on various approaches, evaluating the results with a linear classifier (baseline) and with our KAC. It demonstrates that our KAC consistently improves their performance, especially in long-sequence tasks. The change is indicated next to the accuracy, with blue representing a decrease and red representing an improvement. }
    \label{tab:imagenetR}
    \begin{tabular}{cllllllll}
    \toprule
    \multirow{2}{*}{Method} & \multicolumn{2}{c}{5 steps} & \multicolumn{2}{c}{10 steps} & \multicolumn{2}{c}{20 steps} & \multicolumn{2}{c}{40 steps} \\
    \cmidrule{2-9} 
         & \multicolumn{1}{c}{Avg} &  \multicolumn{1}{c}{Last} &  \multicolumn{1}{c}{Avg} &  \multicolumn{1}{c}{Last} & \multicolumn{1}{c}{Avg} &  \multicolumn{1}{c}{Last} &  \multicolumn{1}{c}{Avg} &  \multicolumn{1}{c}{Last}  \\
    \midrule
    L2P         & 78.42 & 73.57 & 79.58 & 73.10 & 77.93 & 70.35 & 74.28 & 66.02\\
    w/ KAC        & 77.98\lessval{-0.44} & 73.56\lessval{-0.01} & 79.22\lessval{-0.36} & 73.14\changeval{+0.04} & 78.94\changeval{+1.01} & 72.11\changeval{+1.76} & 76.34\changeval{+2.06} & 69.74\changeval{+3.72} \\
    \midrule
    DualPrompt  & 79.75 & 74.57 & 79.50 & 72.48 & 78.35 & 70.68 & 74.51 & 66.31\\
    w/ KAC        & 79.96\changeval{+0.21} & 76.37\changeval{+1.80} & 80.72\changeval{+1.22} & 75.67\changeval{+3.19} & 80.40\changeval{+2.05} & 74.68\changeval{+4.00} & 76.87\changeval{+2.36} & 71.24\changeval{+4.93} \\
    \midrule
    CODAPrompt  & 82.27 & 77.62 & 82.49 & 77.01 & 80.92 & 74.40 & 76.80 & 69.34\\
    w/ KAC        & 83.75\changeval{+1.48} & 80.14\changeval{+2.52} & 84.43\changeval{+1.94} & 79.24\changeval{+2.23} & 83.59\changeval{+2.67} & 77.94\changeval{+3.54} & 79.79\changeval{+2.99} & 74.31\changeval{+4.97} \\
    \midrule
    CPrompt     & 84.07 & 78.68 & 83.13 & 76.80 & 81.83 & 74.32 & 78.98 & 70.07\\
    w/ KAC        & 84.51\changeval{+0.44} & 79.08\changeval{+0.40} & 83.97\changeval{+0.84} & 78.07\changeval{+1.27} & 82.56\changeval{+0.73} & 75.73\changeval{+1.41} & 80.89\changeval{+1.91} & 72.05\changeval{+1.98} \\
    \bottomrule
    \end{tabular}
\end{table*}

\subsection{Kolmogorov-Arnold Classifier for CIL}
\label{sec:KAC}
%上述分析证明了KAN layer with RBF 能够benifit CIL，这motivate us to present our Kolmogorov-Arnold Classifier for CIL, which can plug in any CIL approach replacing the conventional linear classifier.
The above analysis demonstrates that the KAN layer with RBF can benefit CIL, motivating us to introduce our Kolmogorov-Arnold Classifier (KAC), which can be integrated into any CIL approach by replacing the conventional linear classifier with it.

An overview of the KAC is shown in Fig.~\ref{fig:pipeline}. The KAC firstly regularizes the feature distribution with a Layer Normalization $\mathcal{LN}$, resulting in a normalized embedding $\mathcal{LN}(F(\bm{x})) = [x_1^{\prime}, x_2^{\prime}, \cdots, x_n^{\prime}]$. After that, it incorporates a KAN layer that includes $N$ Gaussian Radial Basis Functions centered at $c = [c_1, c_2, \cdots, c_N]$. With the basis function $\phi$ is like defined in eq.~\ref{eq:theta}, the logit $l$ is then calculated as:

\begin{equation}
        l = \mathrm{KAC}\big(F(\bm x)\big) = \mathrm{diag}\bigg(W_C\cdot 
        \Phi\Big(\mathcal{LN}\big(F(\bm{x})\big)\Big)\cdot W_{q}\bigg),
\end{equation}

\noindent where $\mathrm{diag}(.)$ represents extracting the diagonal elements of a matrix and the $\Phi\Big(\mathcal{LN}\big(F(\bm{x})\big)\Big)$ is the learnable Gaussian RBF and it can be calculated like:

\begin{equation}
    % \begin{aligned}
    % &\Phi\Big(\mathcal{LN}\big(F(x)\big)\Big) =\\ 
        \begin{pmatrix}
            \phi (||x_1^{\prime} - c_1||)& \phi (||x_1^{\prime} - c_2||)& \cdots& \phi (||x_1^{\prime} - c_N||)\\
            \phi (||x_2^{\prime} - c_1||)& \phi (||x_2^{\prime} - c_2||)& \cdots& \phi (||x_2^{\prime} - c_N||)\\
            \vdots & \vdots & \ddots & \vdots \\
            \phi (||x_n^{\prime} - c_1||)& \phi (||x_n^{\prime} - c_2||)& \cdots& \phi (||x_n^{\prime} - c_N||)\\            
        \end{pmatrix},
    % \end{aligned}
\end{equation}

\noindent in which $n$ is the dimensionality of the input embedding and $W_C \in \mathbb{R}^{C\times n}$ is a learnable weight matrix that serves as an output linear function to predict the probability for each class, corresponding to the $\Phi_p$ in conventional KAN, while the $W_{q}\in \mathbb{R}^{N\times C}$ corresponds to the $\phi_{p,q}$ in conventional KAN to serve as the univariate learnable activation for each channel for every class. In practice, the $W_C$ and $W_{q}$ can be consolidated into a single weight matrix $W\in \mathbb{R}^{C\times(N\times n)}$, from which the final logit is directly predicted using the basis functions $\phi$. The KAC is then represented as:
\begin{equation}
    \mathrm{KAC}\big(F(\bm x)\big) = W\cdot \mathrm{Reshape}\bigg(\Phi\Big(\mathcal{LN}\big(F(\bm x)\big)\Big)\bigg).
\end{equation}
The $\mathrm{reshape}(.)$ function flattens the $N\times n$ matrix into a 1D vector to facilitate calculations with $W$. 

In a CIL scenario, $T$ tasks arrive sequentially with class counts $[C_1, C_2, \cdots, C_T]$. KAC expands $W$ to accommodate new classes, similar to conventional classifiers~\citep{smith2023coda}. At the $t$th step, there is an old classification matrix $W^{t-1} \in \mathbb{R}^{(N\times n) \times C_{old}}$, where $C_{old} = C_1 + C_2 + \cdots + C_{t-1}$, and a new matrix $W^t \in \mathbb{R}^{(N\times n) \times C_t}$, with the final $W$ after the $t$th step being the concatenation of these two matrices.

\section{Experiments}
\subsection{Benchmarks \& Implementations}

\myPara{Benchmarks.}
We evaluate the CIL scenario and further validate the robustness of our method in Domain Incremental Learning (DIL)~\citep{wang2022s}. For CIL, we conduct experiments on two commonly used datasets, ImageNet-R~\citep{hendrycks2021many} and CUB200~\citep{wah2011caltech}, each containing 200 classes. Starting with 0 base classes, all classes are separated into 5, 10, 20, and 40 steps to feed the model for training sequentially. 
%对于 DIL, Following Sprompt, 我们将domainnet dataset split into 6 domains, in which 345 categories are classified.
For DIL, following Sprompt~\citep{wang2022s}, we split the DomainNet~\citep{peng2019moment} dataset into 6 domains, classifying a total of 345 categories across all tasks. All experiments are conducted in a non-exemplar setting, with no old samples saved for new training. The results of experiments with various seeds are presented in the supplementary materials.

\begin{table*}[tbp]
    \centering
    \setlength{\tabcolsep}{6.0pt}
    \small
    \caption{Results on CUB200 dataset. The average incremental accuracy and the last accuracy are reported. KAC delivers significant improvements for all baselines, especially in long-sequence tasks, highlighting its superior performance on fine-grained datasets.}
    \label{tab:cub}
    \begin{tabular}{cllllllll}
    \toprule
    \multirow{2}{*}{Method} & \multicolumn{2}{c}{5 steps} & \multicolumn{2}{c}{10 steps} & \multicolumn{2}{c}{20 steps} & \multicolumn{2}{c}{40 steps} \\
    \cmidrule{2-9} 
         & \multicolumn{1}{c}{Avg} &  \multicolumn{1}{c}{Last} &  \multicolumn{1}{c}{Avg} &  \multicolumn{1}{c}{Last} & \multicolumn{1}{c}{Avg} &  \multicolumn{1}{c}{Last} &  \multicolumn{1}{c}{Avg} &  \multicolumn{1}{c}{Last}  \\
    \midrule
    L2P         & 80.05 & 76.04 & 74.02 & 65.28 & 63.31 & 51.78 & 46.84 & 35.41 \\
    w/ KAC        & 84.42\changeval{+4.37} & 83.80\changeval{+7.76} & 81.54\changeval{+7.52} & 79.77\changeval{+14.49} & 73.70\changeval{+10.39} & 70.13\changeval{+18.35} & 66.08\changeval{+19.24} & 60.43\changeval{+25.02} \\
    \midrule
    DualPrompt  & 81.84 & 76.38 & 75.10 & 64.60 & 66.89 & 54.68 & 50.61 & 37.55 \\
    w/ KAC        & 86.20\changeval{+4.36} & 85.03\changeval{+8.65} & 82.18\changeval{+7.08} & 79.61\changeval{+14.01} & 76.93\changeval{+10.04} & 71.91\changeval{+17.23} & 71.31\changeval{+20.70} & 64.69\changeval{+27.14} \\
    \midrule
    CODAPrompt  & 83.09 & 78.73 & 79.30 & 71.87 & 69.49 & 58.00 & 52.57 & 37.81 \\
    w/ KAC        & 86.56\changeval{+3.47} & 85.61\changeval{+6.88} & 85.04\changeval{+5.74} & 82.59\changeval{+10.72} & 77.23\changeval{+7.74} & 73.32\changeval{+15.32} & 71.36\changeval{+18.79} & 64.56\changeval{+26.75} \\
    \midrule
    CPrompt     & 88.62 & 82.02 & 85.77 & 76.80 & 83.97 & 72.99 & 77.34 & 64.80 \\
    w/ KAC        & 89.60\changeval{+0.98} & 83.08\changeval{+1.06} & 89.04\changeval{+3.27} & 80.75\changeval{+3.95} & 87.06\changeval{+3.09} & 78.54\changeval{+5.55} & 85.11\changeval{+7.77} & 76.51\changeval{+11.71} \\
    \bottomrule
    \end{tabular}
\end{table*}

\myPara{Implementation Details.}
All experiments are conducted with ViT-B/16 backbones. The numbed of RBFs is set to 4, the centers $[c_1, c_2, \cdots, c_N]$ are evenly distributed between -2 and 2, and the $\sigma$ in the Gaussian functions is set to 1, allowing for an average division of the range.
To validate the effectiveness of KAC, we select four prompt-based CIL approaches L2P~\citep{wang2022learning}, DualPrompt~\citep{wang2022dualprompt}, CODAPrompt~\citep{smith2023coda} and CPrompt~\citep{gao2024consistent} as baselines, all of which have achieved superior performance across various CIL benchmarks. These approaches leverage learnable prompts to extract information from pre-trained backbones and classify the extracted embeddings using linear classifiers. We directly replace their classifiers with KAC with their original hyperparameters to train the model, allowing for a comparison of the differences between classifiers. We implement all compared approaches with their official code and their original selected hyperparameters.
For all experiments, we report the average incremental accuracy (the average accuracy over all tasks) and the accuracy of the last task (the overall accuracy after learning the final task).

\subsection{Experimental Results}

\myPara{Experiments on ImageNet-R.}
Tab.~\ref{tab:imagenetR} compares the accuracies between the baseline methods and those with KAC in the ImageNet-R benchmarks. Replacing the linear classifiers with KAC leads to improvements across all methods, especially in challenging long-sequence scenarios, where gains of 3 to 5 points are observed in most cases. It demonstrates that KAC effectively helps models mitigate forgetting at each step. 
%此外，对比CODAPrompt和CPrompt我们发现，尽管两者在应用线性分类器时效果不相上下，但是换成KAC后CODAPrompt明显超过了CPrompt，这说明KAC与不同方法的兼容性是不同的。
Furthermore, comparing CODAPrompt and CPrompt, we find that while both perform similarly when using linear classifiers, CODAPrompt outperforms CPrompt when switched to KAC. This indicates that the compatibility of KAC with different methods varies.

\myPara{Experiments on CUB200.}
Tab.~\ref{tab:cub} shows a comparison of the metrics in the CUB200 settings, surprising improvements achieving 10 to 25 percent are observed in long-sequence scenarios. As CUB200 is a fine-grained bird classification dataset, we believe that KAC will perform well with such fine-grained datasets.

\myPara{Experiments on DomainNet.}
%与前两个setting不同，我们在DomainNet上进行domain Incremental learning的实验，该实验旨在证明我们的KAC的鲁棒性，具有拓展到任意continual classification task的能力。
%
We conduct experiments on DomainNet for Domain Incremental Learning, aiming to validate the ability of KAC to extend to other continual classification tasks. As shown in Tab.~\ref{tab:domainnet}, when all approaches are implemented with KAC, the performance achieves an improvement of about 1 percent in average incremental accuracy and about 0.5 percent in last accuracy, demonstrating the robustness of our KAC.

\begin{figure}[tp!]
    % \begin{minipage}{.48\linewidth}
    \centering
    \setlength{\abovecaptionskip}{0pt}
    % \vspace{-10pt} % 减少图片上方的间距
    \includegraphics[width=0.9\linewidth]{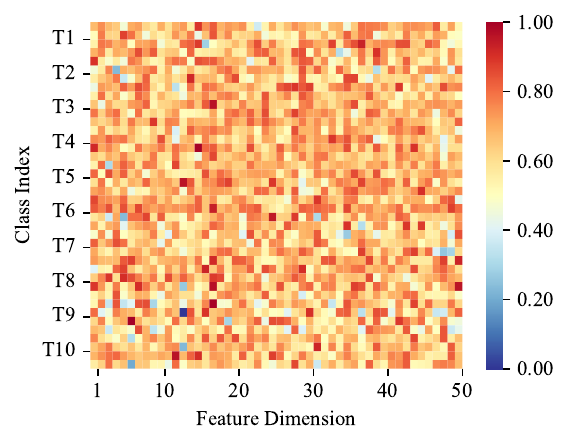}
    % \vspace{-18pt} % 减少图片下方的间距
    %不同类别对不同channel的activation maps. x-axis is 50 random selected channels, while y-axis represents classes in different tasks.
    \caption{Activation maps for different classes across different channels. The x-axis represents 50 randomly selected channels from feature embeddings, while the y-axis represents classes from different tasks. The colors indicate varying levels of interest.}
    \label{fig:attn}
\end{figure}

\myPara{Visualization of activation maps.}
Fig.~\ref{fig:attn} illustrates how different classes activate distinct channels, the differences in activation across different channels for various classes. Only a subset of channels is activated for each category, and updates are applied exclusively to these channels, preventing any impact on the other channels and highlighting the locality advantage in mitigating catastrophic forgetting.

\subsection{Ablation Study and Analysis}

\begin{figure}
    \centering
        \includegraphics[width=.8\columnwidth]{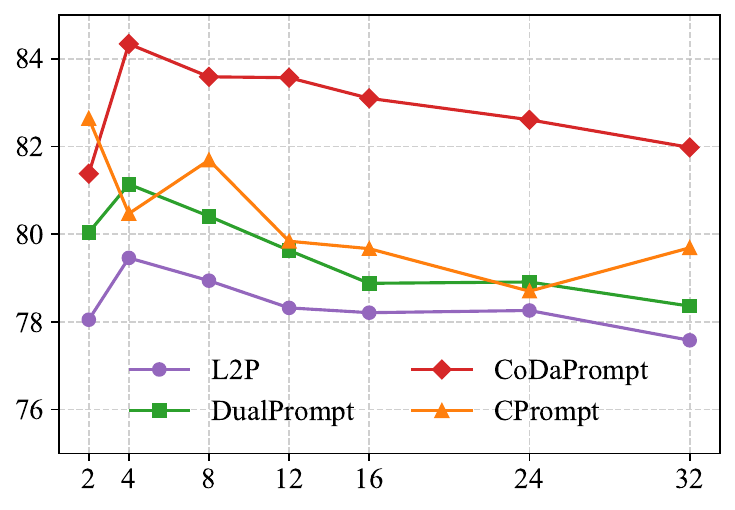}
        \caption{Ablation study on different numbers of basis functions in the 20 steps ImageNet-R scenario. The x-axis represents the number of basis functions, while the y-axis indicates the average incremental accuracy with varying numbers.}
        \label{fig: numgrids}
\end{figure}

\myPara{Ablation on the number of basis functions.}
The number of basis functions $N$ is a key hyperparameter of KAC. An excessive number of basis functions may lead to additional computations and result in a significantly high dimensionality of $W$. Conversely, a small value of $N$ may compromise the approximation capacity of KAC. To explore an appropriate value for $N$, we conduct an ablation study on it. Fig.~\ref{fig: numgrids} shows the average incremental accuracy for four approaches using KAC with different numbers of basis functions in the 20 steps experiment on ImageNet-R. The results indicate that simply increasing the number of basis functions does not benefit mitigating forgetting, and further demonstrate that the performance improvement is not due to  increasing the dimensionality of the embedding space. Most approaches exhibit better performance when $N = 4$ or $N = 8$, encouraging us to set $N$ as 4 in our experiments.

\begin{table}[t]
        \centering
        %我们将KAC当中的spline functions替换为MLP来验证KAN structure的有效性。其中，wmlp和wmlp（fixed）分别代表MLP随模型一起训练，以及随机初始化的MLP projection。实验在20steps imagenetr scenario进行
        \small
    \centering
    \setlength{\tabcolsep}{18.0pt}
    \captionof{table}{Results on DomainNet. A Domain Incremental Learning experiment is conducted on it with 6 incremental domains of 345 classes. We report the average incremental accuracy and the accuracy of the last task. The results show that KAC can also work in DIL settings.}
    \label{tab:domainnet}
    \begin{tabular}{cll}
    \toprule
        \multirow{1}{*}{Method}       
        % & \multicolumn{2}{c}{} \\
        % \cmidrule{2-3}
        & \makecell[c]{Avg}   & \makecell[c]{Last} \\
    \midrule
        L2P          & 57.78  & 49.22 \\
        w/ KAC& 59.79\changeval{+2.01} & 51.10\changeval{+1.88} \\ \midrule
        DualPrompt   & 60.96  & 51.83\\  
        w/ KAC& 62.06\changeval{+1.10} & 52.76\changeval{+0.93} \\ \midrule
        CODAPrompt   & 61.61  & 53.12  \\
        w/ KAC& 62.78\changeval{+1.17} & 53.54\changeval{+0.42} \\ \midrule
        CPrompt   & 61.32  & 52.49  \\
        w/ KAC& 62.13\changeval{+0.81} & 53.02\changeval{+0.53} \\
    \bottomrule
    \end{tabular}
\end{table}

\begin{table}[t]
    % \begin{minipage}{0.5\linewidth}
        \centering
        \small
        %我们将KAC当中的spline functions替换为MLP来验证KAN structure的有效性。其中，wmlp和wmlp（fixed）分别代表MLP随模型一起训练，以及随机初始化的MLP projection。实验在20steps imagenetr scenario进行
        \caption{Ablation study on the structure of the classifier. We replace the spline functions in KAC with MLPs to validate the effectiveness of the KAN structure. Here, w/ MLP represents the MLP trained alongside the model, while w/ MLP (fixed) represents the randomly initialized MLP projection without any updating. The experiments are conducted in the 20 steps ImageNet-R scenario.}
        \label{tab:structure}
        \begin{tabular}{ccccc}
        \toprule
        %              & Avg   & Last  \\ \midrule
        % CODAPrompt     & 80.92 & 74.40 \\
        % $w$ KAC        & 83.59 & 77.94 \\
        % % $w$ KAN$\times$2        & 83.59 & 77.94 \\
        % $w$ MLP        & 80.56 & 73.59 \\
        % $w$ MLP (fixed) & 65.87 & 51.03\\
             & CODAPrompt  & w/ KAC & w/ MLP & w/ MLP (fixed) \\ \midrule
        Avg  & 80.92 & 83.59 & 80.56 & 65.87 \\
        Last & 74.40 & 77.94 & 73.59 & 51.03 \\
        \bottomrule
        \end{tabular}
\end{table}

\myPara{The KAN structure plays a key role.}
To demonstrate that the advantages of KAC lie in the introduced KAN structure, not the additional computations, we replace the RBFs with an MLP layer, setting its output dimension to the number of classes and hidden dimension to $N \times n$ to align the number of parameters with KAC using RBFs, allowing us to make a fair comparison between the two structure. Tab.~\ref{tab:structure} shows the performance of replacing RBFs with the conventional linear classifier with an additional MLP structure implemented on CODAPrompt. Upon comparison, we discover that whether the additional MLP structure is updated alongside the model or not, it does not yield any positive effects. This indicates that the advantages of KAC stem from its KAN structure rather than a simple increase in the dimensionality of the classification space.

\myPara{Efficiency analysis.}
In comparison to conventional linear classifiers, our KAC introduces a negligible increase in computational cost and parameter count at the classifier layer. KAC applies fixed Gaussian activation functions to each dimension which almost introduces no extra computations.
For a ViT network with an embedding dimension of 768 to classify 100 categories, the additional parameters introduced by KAC amount to only 0.23M, which is negligible compared to 86M parameters of the backbone.
% \begin{figure*}

    % \end{minipage}
    % \begin{minipage}{.5\linewidth}

% \end{figure*}
\section{Conclusions}

In this paper, we explore the application of Kolmogorov-Arnold Networks (KAN) in continual learning and propose a novel continual classifier, the Kolmogorov-Arnold Classifier (KAC), which leverages KAN's inherent locality capability to mitigate feature shifts during the learning of new tasks. Our analysis reveals that the limited approximation ability of the B-spline functions in KAN, when applied to high-dimensional data, forces the model backbone to introduce more shifts to accommodate new classes, leading to significant degradation in continual learning performance. This exacerbates the model’s forgetting, overshadowing the benefits of locality capability, compared to a traditional linear classifier. To address this issue, we replace the B-spline functions in KAN with Radial Basis Functions (RBFs), which improves performance. KAC demonstrates substantial advantages across various continual learning scenarios, underscoring its effectiveness and robustness. In the future, we plan to explore further possibilities of KAN in continual learning, fully harnessing its inherent strengths.

\section{Acknowledgments}
This work was funded by  
NSFC (NO. 62206135, 62225604), Young Elite Scientists Sponsorship Program by CAST (2023QNRC001), “Science and Technology Yongjiang 2035" key technology breakthrough plan project (2024Z120), Shenzhen Science and Technology Program (JCYJ20240813114237048), and the Fundamental Research Funds for the Central Universities 
(Nankai Universitiy, 070-63233085). 
Computation was supported by the Supercomputing Center of Nankai University.

{
    \small
    \bibliographystyle{ieeenat_fullname}
    \bibliography{main}
}

% rag baseline
% rag插件，继承了rag的模型

\clearpage
\setcounter{page}{1}
\maketitlesupplementary

\section{More Experimental Results}

\paragraph{Experiments on CIFAR-100.} Tab.~\ref{tab:cifar100} compares the average incremental accuracy between the baseline methods and those with KAC in the CIFAR-100 dataset. Replacing the linear classifiers with KAC improves most of the methods, with a little drop in CPrompt and L2P. Due to the low pixel resolution of CIFAR-100, it is generally suitable for training smaller-scale networks. For pre-trained backbones, performance tends to be saturated, which is why our method does not show significant improvement on this dataset.
\begin{table}[htbp]
    \centering
    \caption{The average incremental accuracy of CIFAR-100 10 steps scenario.}
    \label{tab:cifar100}
    \begin{tabular}{lcc}
    \toprule
        Method       & Linear    & KAC               \\
    \midrule
        L2P          & 83.78             & 83.71 \changeval{(-0.07)} \\
    \midrule
        DualPrompt   & 84.80             & 85.74 \changeval{(+0.94)} \\
    \midrule
        CODAPrompt   & 86.65             & 87.26 \changeval{(+0.61)} \\
    \midrule
        CPrompt      & 87.50             & 87.19 \changeval{(-0.31)} \\
    \bottomrule
    \end{tabular}
\end{table}

\begin{table}[htbp]
    \centering
    \caption{Comparison of the KAC and KAC with shortcut on the last accuracy in ImageNet-R 20 steps scenario.}
    \label{tab:ab_shortcut}
    \begin{tabular}{lccc}
    \toprule
        Method      & Baseline & KAC + Shortcut & KAC \\
    \midrule
        L2P         & 70.35    & 71.09           & 72.11 \\
        DualPrompt  & 70.68    & 72.83           & 74.68 \\
        CODAPrompt  & 74.40    & 75.57           & 77.94 \\
        CPrompt     & 74.32    & 73.55           & 75.73 \\
    \bottomrule
    \end{tabular}
\end{table}
\paragraph{Results of multiple runs on CUB200.}
To further evaluate the robustness of KAC, we generate random class sequences using multiple seeds and evaluate the performance of various methods using KAC and conventional linear classifiers. Specifically, we report the last accuracy for each setting on CUB200. Table~\ref{tab:cub_repeat} reports the mean and standard deviation of these metrics. The results demonstrate that KAC consistently outperforms the baseline methods across most experiments.

\begin{table*}[tbp]
    \centering
    \setlength{\tabcolsep}{10.0pt}
    \caption{The results of multiple runs on CUB200 dataset with more seeds. The mean and std of the last accuracy are reported.}
    \label{tab:cub_repeat}
    \begin{tabular}{lcccccccc}
    \toprule
    \multirow{2}{*}{Method} & \multicolumn{2}{c}{5 steps} & \multicolumn{2}{c}{10 steps} & \multicolumn{2}{c}{20 steps} & \multicolumn{2}{c}{40 steps} \\
    \cmidrule{2-9} 
         & \multicolumn{1}{c}{mean} &  \multicolumn{1}{c}{std} &  \multicolumn{1}{c}{mean} &  \multicolumn{1}{c}{std} & \multicolumn{1}{c}{mean} &  \multicolumn{1}{c}{std} &  \multicolumn{1}{c}{mean} &  \multicolumn{1}{c}{std}  \\
    \midrule
    L2P         & 76.60 & 0.79 & 69.23 & 2.86 & 59.11 & 5.06 & 41.65 & 4.71 \\
    $w$ KAC        & 80.35 & 2.60 & 77.07 & 1.92 & 73.32 & 0.57 & 65.89 & 4.09 \\
    \midrule
    DualPrompt  & 77.35 & 1.68 & 71.13 & 1.88 & 61.91 & 5.03 & 44.98 & 5.21 \\
    $w$ KAC        & 82.39 & 1.95 & 79.55 & 1.88 & 75.94 & 1.76 & 69.86 & 4.28 \\
    \midrule
    CODAPrompt  & 75.62 & 2.65 & 70.77 & 0.71 & 62.58 & 3.44 & 45.14 & 5.49 \\
    $w$ KAC        & 82.88 & 1.96 & 78.74 & 2.73 & 74.74 & 0.194 & 70.57 & 4.27 \\
    \bottomrule
    \end{tabular}
\end{table*}

\section{More Ablation Studies}
\textbf{Ablation on the linear shortcut.} In KAC, we don't follow conventional KAN, in which a linear shortcut is added with the spline functions. In this section, we show that the linear shortcut cannot help KAC achieve better performance. Tab.~\ref{tab:ab_shortcut} reports the accuracy of the last task in ImageNet-R 20 steps scenario. It demonstrates that when linear shortcut is added, it achieves even worse accuracy, supporting our decision to remove the linear shortcut.

\section{More Visualizations}
\begin{figure*}[htbp]
    \centering
    \begin{subfigure}[b]{0.24\textwidth}
        \centering
        \includegraphics[width=\linewidth]{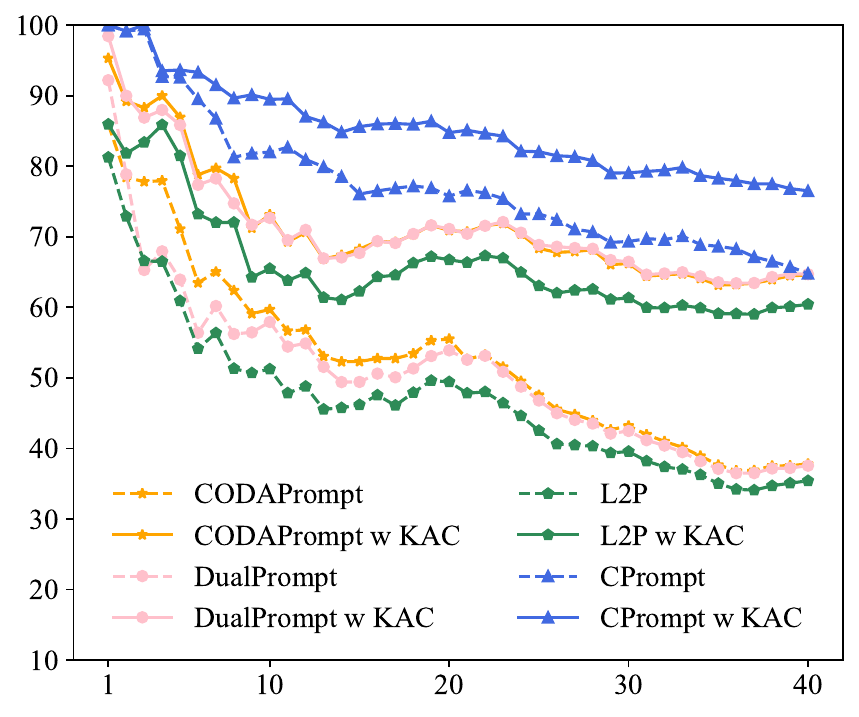}
        \caption{40 steps}
        \label{fig:image1}
    \end{subfigure}
    % \hspace{0.05\textwidth} % 调整图像间距
    \begin{subfigure}[b]{0.24\textwidth}
        \centering
        \includegraphics[width=\linewidth]{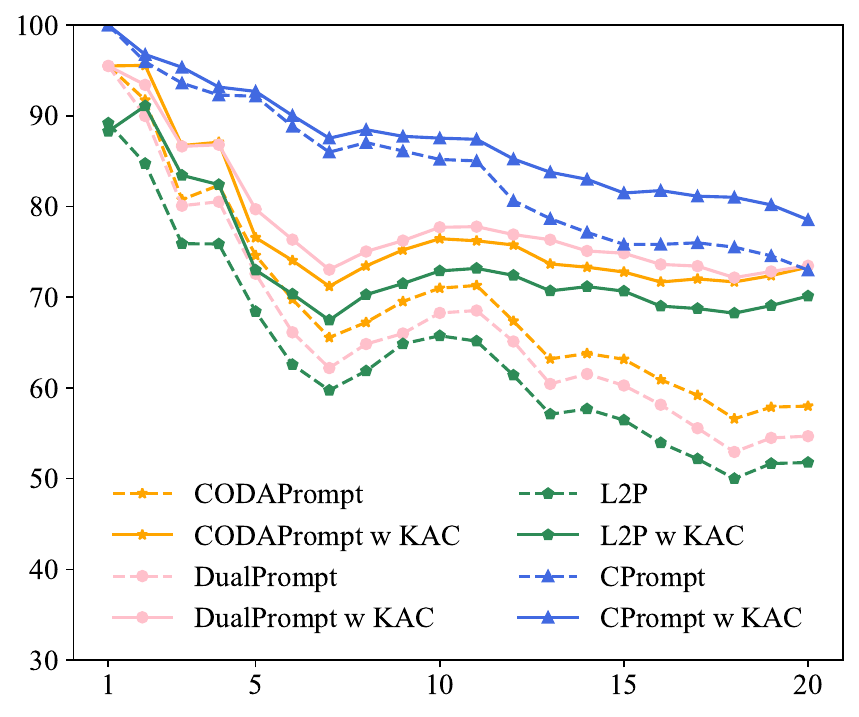}
        \caption{20 steps}
        \label{fig:image2}
    \end{subfigure}
    % \vspace{0.05\textheight} % 调整行间距
    \begin{subfigure}[b]{0.24\textwidth}
        \centering
        \includegraphics[width=\linewidth]{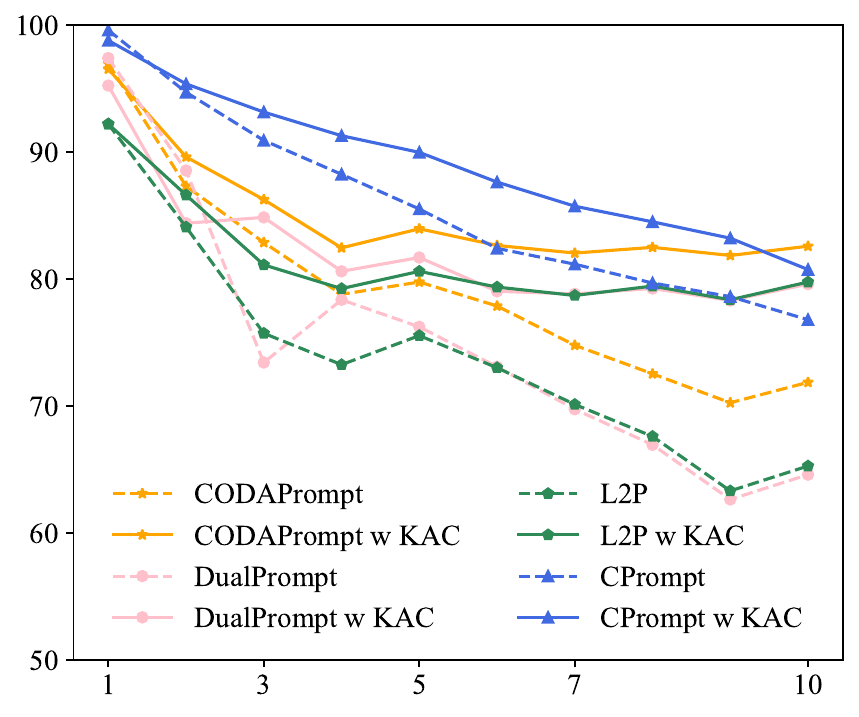}
        \caption{10 steps}
        \label{fig:image3}
    \end{subfigure}
    % \hspace{0.05\textwidth}
    \begin{subfigure}[b]{0.24\textwidth}
        \centering
        \includegraphics[width=\linewidth]{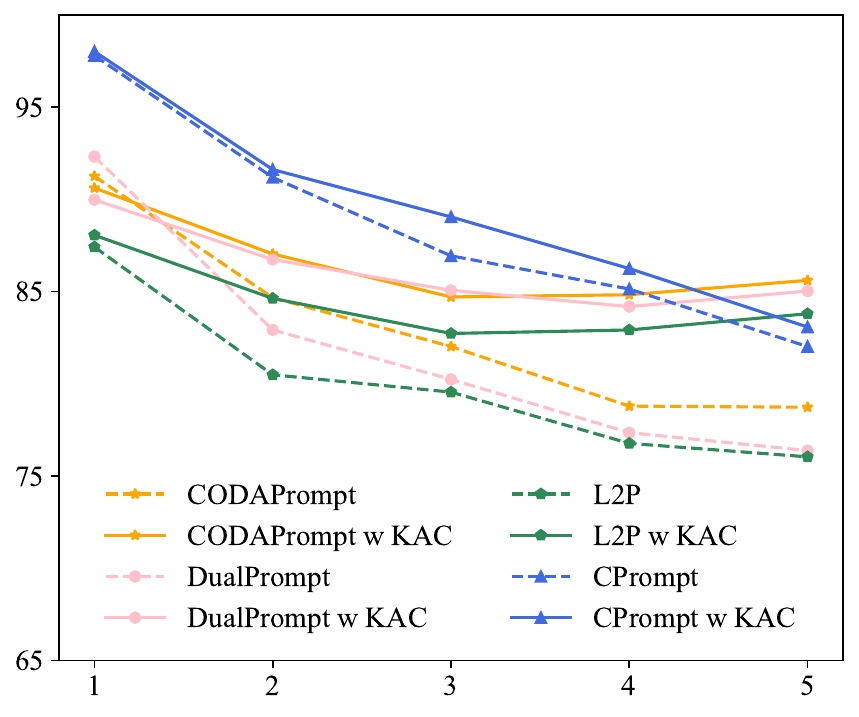}
        \caption{5 steps}
        \label{fig:image4}
    \end{subfigure}
    %在CUB200数据集上不同steps 的scenarios的精度曲线，横坐标代表逐渐增加的任务，纵坐标代表精度，可以看到KAC于baseline的变化趋势相同，但是在每一步的遗忘更小些。
    \caption{The accuracy curves for scenarios of different steps on the CUB200 dataset. The x-axis represents the gradually increasing tasks and the y-axis represents accuracy at each step. It can be observed that KAC follows the same trend as the baseline, but exhibits less forgetting at each step.}
    \label{fig:visual_cub}
\end{figure*}

\textbf{Visualization of performance on CUB200.} 
%为了探究KAC在CUB200上性能优异的原因，我们绘制并观察了CUB200在不同steps的实验当中的精度曲线。如图6所示，随着任务数量的不断增加，KAC表现出越来越大的优势，其中，在一些步骤当中，baseline经常会产生剧烈的下降，而在这些步骤中KAC往往展现出比linear classifier更小的遗忘，这帮助KAC积累最终得到较高的精度。
To investigate the reasons behind the superior performance of KAC on CUB200, we make an observation on the accuracy curves of CUB200 across experiments with different steps. As shown in Fig.~\ref{fig:visual_cub}, with the arriving of tasks, KAC demonstrates a growing advantage. In several steps, the baseline frequently experiences significant forgetting, while KAC often exhibits less forgetting compared to the linear classifier during these steps, which helps KAC accumulate a higher final accuracy.

\end{document}